\documentclass[letterpaper, 10 pt, journal]{ieeetran}

\usepackage{times}
\usepackage{epsfig}
\usepackage{graphicx}
\usepackage{amsmath}

\usepackage{amssymb}
\usepackage{amsfonts}
\usepackage{wrapfig}
\usepackage{cite}
\usepackage{color,soul}
\usepackage[ruled,vlined]{algorithm2e}
\usepackage{lipsum}
\usepackage{cuted}
\usepackage{mathrsfs }
\usepackage{mathtools}
\usepackage{algorithmic}
\usepackage[ruled,vlined]{algorithm2e}
\usepackage{enumerate}
\usepackage{makeidx}  
\usepackage{bm}
\usepackage{verbatim}
\usepackage{subfigure}
\usepackage{multirow}
\usepackage{float}
\usepackage{booktabs}

\newtheorem{Thm}{Theorem}
\newtheorem{Def}{Definition}

\begin{document}

\title{Model Quality Aware RANSAC: A Robust Camera Motion Estimator
\author{Shu-Hao~Yeh and Dezhen~Song
\thanks{*This work was supported in part by the National Science Foundation under NRI-1526200, 1748161, and 1925037.}
\thanks{S. Yeh and D. Song are with CSE Department, Texas A\&M University, College Station, TX 77843, USA. Emails: \textit{ericex1015@tamu.edu} and \textit{dzsong@cs.tamu.edu}.}
}
}
\maketitle
\thispagestyle{plain}
\pagestyle{plain}

\begin{abstract}
Robust estimation of camera motion under the presence of outlier noise is a fundamental problem in robotics and computer vision. Despite existing efforts that focus on detecting motion and scene degeneracies, the best existing approach that builds on Random Consensus Sampling (RANSAC) still has non-negligible failure rate. Since a single failure can lead to the failure of the entire visual simultaneous localization and mapping, it is important to further improve robust estimation algorithm. We propose a new robust camera motion estimator (RCME) by incorporating two main changes: model-sample consistence test at model instantiation step and inlier set quality test that verifies model-inlier consistence using differential entropy. We have implemented our RCME algorithm and tested it under many public datasets. The results have shown consistent reduction in failure rate when comparing to RANSAC-based Gold Standard approach. More specifically, the overall failure rate for indoor environments has reduced from $1.41\%$ to $0.02\%$.
\end{abstract}

\section{Introduction}
Robust estimation of geometric relationship between two camera views is a fundamental problem in computer vision and robotics. It simultaneously identifies corresponding inlier features from outlier noises. When applying to the problem of visual odometry (VO) or visual simultaneous localization and mapping (vSLAM) in robotics, the geometric relationship is often the fundamental matrix, or essential matrix when camera intrinsic parameters are known. Since camera motion can be inferred from essential matrix, this is also known as camera motion estimation.

\begin{figure}[ht]
\centering
\subfigure[]{\includegraphics[width = 1.1in, trim={0.5cm 0.5cm 0.2cm 0.5cm},clip]{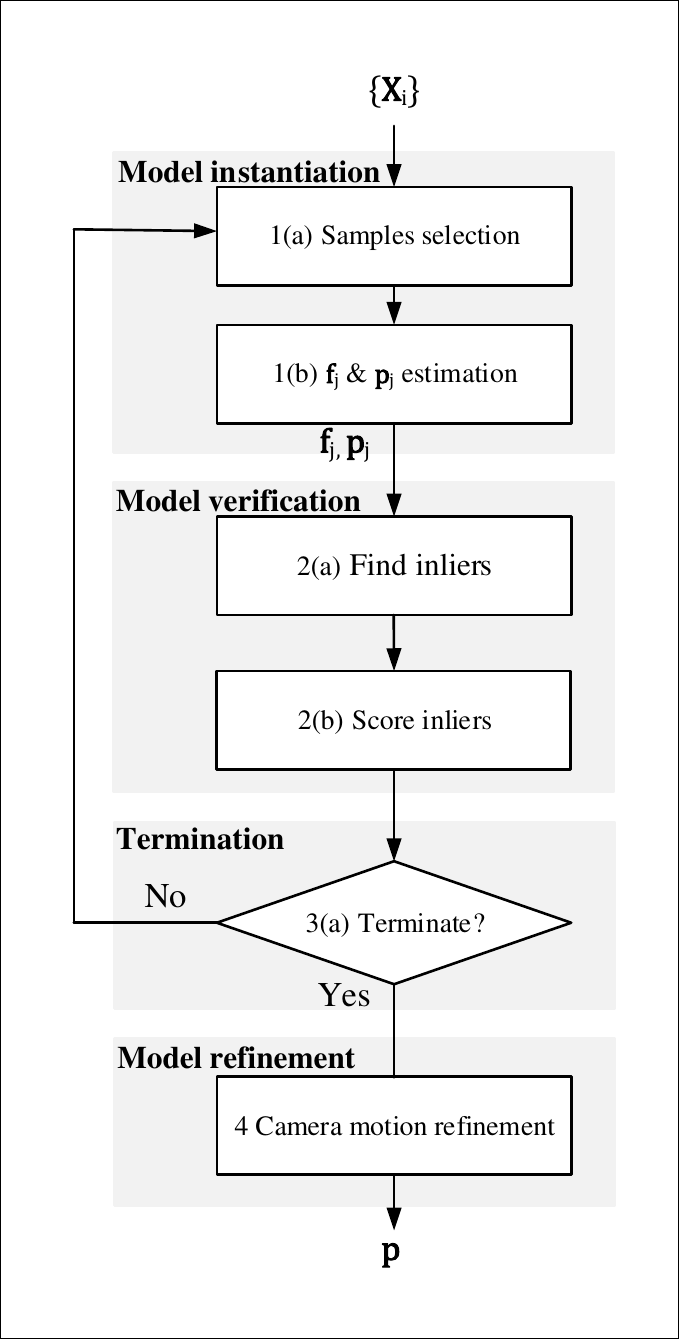}\label{fig::RANSAC_system_diagram}}
\subfigure[]{\includegraphics[width = 2.2in, trim={0.5cm 0.5cm 0.5cm 0.5cm},clip]{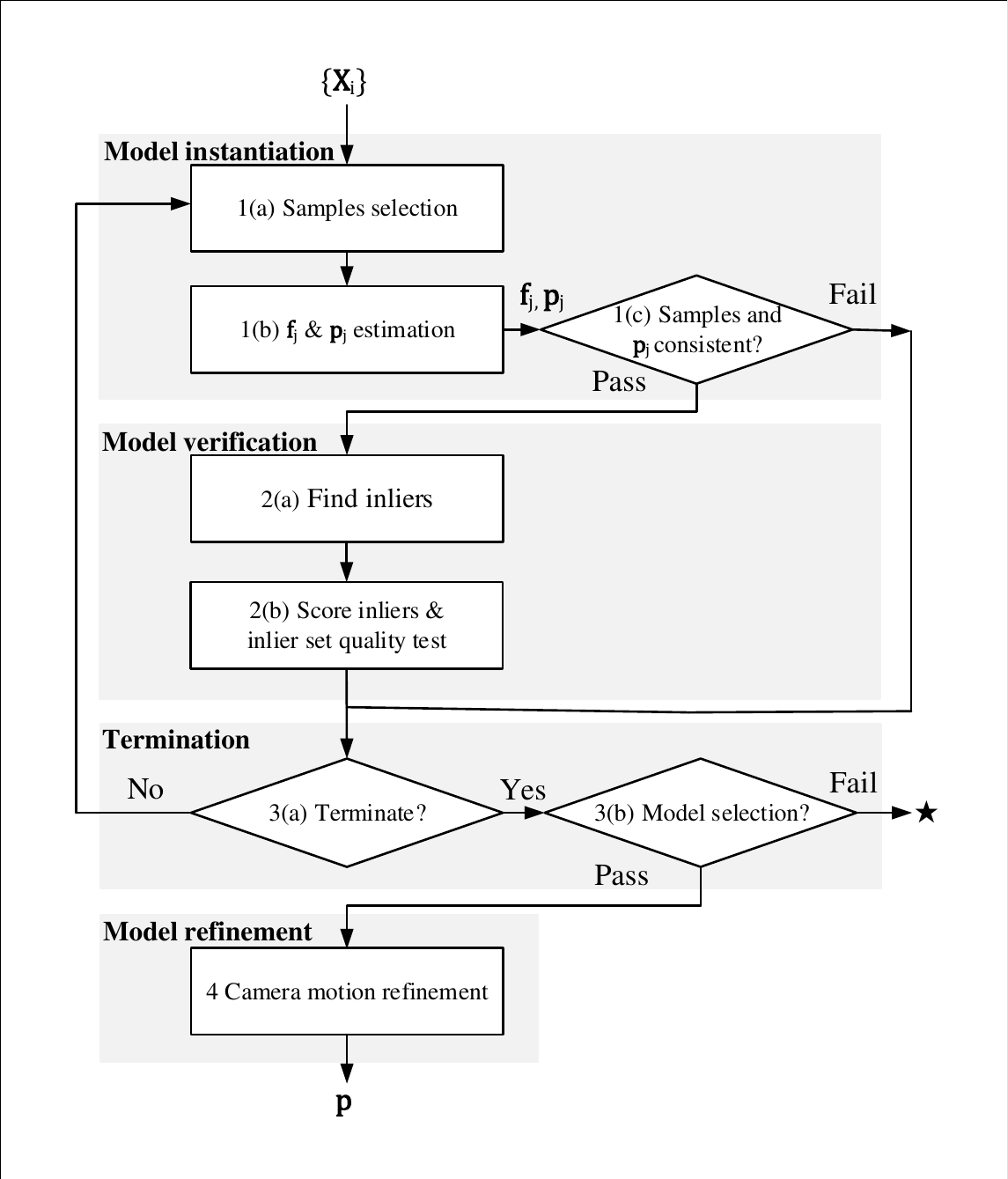}\label{fig::RCME_system_diagram}}
\caption{The system diagrams of our RCME framework in (b) which is an improvement of RANSAC in (a), the Gold standard algorithm for estimating camera motion.}
\label{fig::system_diagram}
\end{figure}

The classical robust estimation method that can filter out outliers is random consensus sampling (RANSAC)~\cite{fischler1981random} (Fig.~\ref{fig::RANSAC_system_diagram}). Simply applying RANSAC to estimate camera motion is not wise since there are well-known motion degeneracy and scene degeneracy issues, also known as Type A and Type B degeneracies according to~\cite{torr1999problem}, respectively. To eliminate both degeneracies, existing state-of-the-art approaches apply key frame selection and simultaneously estimate homography and fundamental matrices~\cite{mur2017orb}. However, these remedies do not solve all the problems. After testing 14816 images from 23 different datasets, our results show that the RANSAC-based fundamental matrix estimation algorithm still has a failure rate of $1.41\%$ in indoor environments, despite that these test data have already both degeneracies removed. Since one single failure can lead to an entire continuous monocular vSLAM failure, this leads to the mean distance between failure is mere 35 meters if the average distance between adjacent key frames is 0.5 meters. This means that current robust estimation of camera motion is not satisfying.

A deep look into the failed cases reveals that due to the fact that the epipolar geometric relationship characterized by a fundamental matrix is a weak point-to-line distance measure and scene feature distribution may not be uniform, it is possible that the initial sampling in RANSAC may establish a wrong model and still can find a potential large number of inliers. As a result, RANSAC may output wrong camera motion. To address these issues, we propose a new robust camera motion estimator (RCME) (see  Fig.~\ref{fig::RCME_system_diagram}) by incorporating two main changes: model-sample consistence test to verify the quality of the model built by over-parameterized sampling, and inlier set quality test that first determines inliers according to model uncertainty and then verifies if the inlier set agrees with the model through a test on differential entropy. The RCME can also detect and report a failed estimation when input data quality is too low.

We have implemented the RCME algorithm in C++ and tested with a variety of public datasets. The results have shown consistent reduction of failure rate across almost all test data. More specifically, the overall failure rate for indoor environments has reduced from $1.41\%$ to $0.02\%$.


\section{Related Work}
Robust camera motion estimation mainly relates to two research fields: robust estimation in computer vision and the front-end algorithm of vSLAM/VO in robotics.

Robust estimation is a fundamental problem in computer vision and robotics. A robust estimator's task is to estimate parameters and find the inliers from noisy data correspondences according to a predefined type of geometric relationship. The outliers from wrong correspondences often introduce significant errors when estimating the geometric relationship. M-estimator, L-estimator and R-estimator~\cite{huber1996robust} formulate the estimation problem as a reprojection error minimization problem and solve it by using a nonlinear cost function. Leat Median of Squares (LMeds) by Rousseeuw~\cite{rousseeuw1984least} minimizes the median of error instead. These methods are not stable when over half of data are outliers. RANSAC by Fischler et al.~\cite{fischler1981random} is the most widely used robust estimator since it is capable of handling a high ratio of outliers.

The vSLAM and VO works can be classified as the feature-based approaches~\cite{davison2007monoslam, klein2007parallel, mur2017orb}, the semi-dense approaches~\cite{engel2013semi, forster2014svo}, and the direct approaches~\cite{engel2014lsd, engel2017direct}. Among these methods, the widely adopted feature-based approaches explicitly and repeatedly utilize robust estimation of camera motion (e.g., MonoSLAM~\cite{davison2007monoslam} and PTAM~\cite{klein2007parallel}). RANSAC has been employed in combination with different fundamental matrix estimation models such as the one-point method in~\cite{scaramuzza20111}, the five-point algorithm~\cite{stewenius2006recent} in PTAM or the eight-point method in ORB-SLAM2 ~\cite{mur2017orb}. These works repeatedly obtain initial solutions for camera poses (key) frame by (key) frame before bundle adjustment. Hence the robustness of the robust estimation method is critical here because a single failure can collapse the entire mapping process which many involve thousands of RANSAC-based camera motion estimation instances.

Our work is an improvement over the existing RANSAC framework for camera motion estimation. In fact, most existing works in this area mainly focus on detecting the two type of degeneracy issues: Type A and Type B~\cite{torr1999problem}. When handling degeneracy cases, existing approaches often treat the problem as a model selection between the homography relationship and the fundamental matrix relationship. Popular approaches include AIC~\cite{akaike1974new}, PLUNDER-DL score~\cite{torr1998robust}, GIC~\cite{kanatani1998geometric}, and GRIC~\cite{torr1998geometric}. Pollefeys et al.~\cite{pollefeys2002surviving} uses GRIC in recovering the structure and motion. ORB-SLAM2~\cite{mur2017orb} initializes the trajectory by finding the better geometry model between the fundamental matrix or the homography matrix. However, even when Type A and Type B degeneracies have been ruled out, our tests have shown that RANSAC-based camera motion estimation may still fail with non-negligible probability. It is due to the fact that the fundamental matrix only represents a weak geometric point-line distance geometric relationship which is not very selective.

Many attempts have been tried to improve RANSAC in general. The loss function  function~\cite{torr2000mlesac} is introduced in model selection. Choi et al.~\cite{choi1997performance} also provides a survey to RANSAC-based methods and categorizes them into three groups: \emph{being fast}, \emph{being accurate}, and \emph{being robust}. Our work emphasizes \emph{being robust}.
Inspired by the unified framework named USAC~\cite{raguram2012usac}, our work employs the information entropy for the hypothesis evaluation to improve model quality and hence increase robustness. The threshold used in determined inliers is based on the model uncertainty. The later step is inspired by Cov-RANSAC~\cite{raguram2009exploiting} which also incorporates the model uncertainty to identify inliers, but they do not rank the model by using the model-inlier consistence.


\section{Problem Definition}

We begin with the following assumptions:
\begin{itemize}
\item[\textbf{a.1}] The camera is pre-calibrated and its lens distortion is removed from images.
\item[\textbf{a.2}] Position noises of the points follow zero-mean Gaussian distribution with known variance $\sigma^{2}$ in each dimension and the noise in each dimension is independent.
\end{itemize}

Let us define the common notations in this paper.
\begin{description}
\item[$\mathrm{K}$] the intrinsic matrix of the camera.
\item[$\mathbf{X}$] is a $4$-vector and concatenates point correspondences from two views. The $i$-th point is denoted as $\mathbf{X}_{i}=\begin{bmatrix}\tilde{\mathbf{x}}^\mathsf{T}_{i}\quad\tilde{\mathbf{x}}^{'\mathsf{T}}_{i}\end{bmatrix}^\mathsf{T}\in\mathbb{R}^{4}$, where $\tilde{\mathbf{x}}_{i}\leftrightarrow \tilde{\mathbf{x}}'_{i}$ is the $i$-th corresponding points from the first and the second views, respectively. Symbol $\tilde{}$ indicates the inhomogeneous coordinate and $\tilde{\mathbf{x}}_{i}, \tilde{\mathbf{x}}'_{i} \in\mathbb{R}^{2}$. The corresponding 3-vector homogeneous representation is $\mathbf{x}_{i}, \mathbf{x}'_{i}\in\mathbb{P}^{2}$.
\item[$\mathbf{f}$] is a $9$-vector with entries from fundamental matrix $\mathrm{F}$, $\mathbf{f} = \begin{bmatrix}\mathrm{F}^{1} & \mathrm{F}^{2} & \mathrm{F}^{3}\\\end{bmatrix}^\mathsf{T}\in\mathbb{R}^{9}$, where $\mathrm{F}^{i}$ denotes the $i$-th row of $\mathrm{F}$. The $i$-th model vector denotes as $\mathbf{f}_{i}$.
\item[$\mathbf{p}$] is a $7$-vector in $\mathrm{SE}(3)$ defining the camera motion and consists of a unit quaternion vector $\mathbf{q}\in\mathbb{R}^{4}$ and a translation vector $\mathbf{t}\in\mathbb{R}^{3}$. The $i$-th camera motion denotes as $\mathbf{p_{i}}=\begin{bmatrix}\mathbf{q}^\mathsf{T}_{i}\quad\mathbf{t}^\mathsf{T}_{i}\end{bmatrix}^\mathsf{T}$.
\end{description}
In this paper, $\mathrm{I}_{n}$ denotes the $n\times n$ identity matrix.

\subsection{Inputs}
We pre-process both views to obtain inputs. Feature detection and feature matching have been applied to obtain putative point correspondences
$\{\mathbf{x}_{i}\leftrightarrow\mathbf{x}'_{i}\}^{n}_{i=1}$, where $n$ is the number of point correspondences. We concatenate every inhomogeneous point correspondence $\tilde{\mathbf{x}}_{i}\leftrightarrow\tilde{\mathbf{x}}'_{i}$ in to a $4$-vector $\mathbf{X}_{i} = \begin{bmatrix}\tilde{\mathbf{x}}^\mathsf{T}_{i}\quad\tilde{\mathbf{x}}^{'\mathsf{T}}_{i}\end{bmatrix}^\mathsf{T}$. The inputs may be obtained from a variety of feature detectors such as scale-invariant feature transform (SIFT)~\cite{lowe2004distinctive}, speeded up robust feature (SURF)~\cite{bay2006surf}, or ORB~\cite{rublee2011orb}. It is worth noting that the point correspondence set is the result of putative matching of feature descriptors and often contains many outliers.

\subsection{Problem Definition}
With notations defined and inputs introduced, our problem is defined as,
\begin{Def}
Given $n$ point correspondences (or points for brevity) $\{\mathbf{X}_{i}\}^{n}_{i=1}$, determine if the solution of the camera motion $\mathbf{p}$ exists. If the solution exists, estimate the camera motion $\mathbf{p}$.
\end{Def}

As shown in the problem definition, one immediate difference between our method and RANSAC is that our method can detect failure cases instead of output an unreliable solution. This is actually very useful for vSLAM because it can be used as a signal to re-adjust key frame selection.

\section{RCME Aglorithm}

\subsection{RANSAC Review and RCME Overview}
Since our RCME is an extension of RANSAC, let us begin with a brief review of RANSAC as shown in Fig.~\ref{fig::RANSAC_system_diagram}. RANSAC establishes hypothesis models by randomly sampling a minimal correspondence set (model instantiation) and then examines each set by comparing how many other features agree with the hypothesis model (model verification). RANSAC is an iterative method which terminates when reaching a maximum iteration number or a high quality inlier set with its size agreeable to the estimated inlier ratio is found. The model with the largest inlier set is selected as the output.

Building on basic RANSAC framework, Fig.~\ref{fig::RCME_system_diagram} illustrates our system diagram. It contains four main blocks in gray. The first three blocks are different from the counterpart. After instantiating a hypothesis model through sampling, we test the consistence between the samples and the model. If no consistence exists between the samples and the model, we discard the model and repeat the model instantiation step. Otherwise, we begin model verification, where we perform an inlier quality test before we select the model according to entropy. If no model can be selected, we consider the camera motion estimation as failure in termination step (See $\star$ from Box 3(b) in Fig.~\ref{fig::RCME_system_diagram}). The selected model is our initial solution for the model refinement.

Without loss of generality, let us assume we are at the $j$-th iteration to begin the explanation.

\subsection{Model Instantiation}\label{sec::model_instantiate}

Model instantiation is the first step of every iteration. We want to check if a model building on the randomly sampled point correspondences actually agrees with the samples. This is crucial for overpameterized models.


Same as the traditional RANSAC, we randomly sample $m$ point-correspondences from $\{\mathbf{X}_{i}\}$ where $m$ is the minimal number of samples to instantiate the model and can be different according to different parameterizations used in modeling. Define the sample set for the $j$-th iteration as
\begin{equation}\label{eq::minimal_sample_set}
    \mathcal{S}_{j}:=\Big\{\mathbf{X}_{s_{k}}:s_{k}\in\{1,\cdots,n\}\quad\text{and}\quad 1\leq k\leq m\Big\},
\end{equation}
where $s_{k}$ is the point index and $k$ denotes the $k$-th sample.

\subsubsection{$\mathbf{f}_{j}\&\mathbf{p}_{j}$ model instantiation and uncertainty analysis}
Given the sample set $\mathcal{S}_{j}$, we instantiate the model $\mathbf{f}_{j}$ and recover the camera motion $\mathbf{p}_{j}$ from $\mathbf{f}_{j}$. Consider a mapping function $\Omega$ which maps the model $\mathbf{f}_{j}$ to the concatenated samples $\begin{bmatrix}\hdots,\mathbf{X}^\mathsf{T}_{s_{k}},\hdots\end{bmatrix}^\mathsf{T}$, $\Omega:\mathbb{R}^{9}\rightarrow\mathbb{R}^{4m}$. The concrete representation of $\Omega$ depends on parameterization of fundamental matrix which includes $1-$point~\cite{scaramuzza20111}, $5-$point~\cite{nister2004efficient}, or $8-$point methods~\cite{Hartley2003}. Since our framework is not limited by a particular parameterization, we represent it as a generic $\Omega$ mapping.

The model instantiation  of fundamental matrix $\mathbf{f}_{j}$ can be represented by
\begin{equation}\label{eq::model_instantiate}
    \mathbf{f}_{j} = \Omega^{-1}([\hdots,\mathbf{X}_{s_{k}}^{\mathsf{T}}, \hdots]^\mathsf{T}).
\end{equation}

The camera motion is recovered from the fundamental matrix decomposition~\cite{hartley2003multiple}.
The camera motion $\mathbf{p}_{j}=\begin{bmatrix}\mathbf{t}^\mathsf{T}_{j}&\mathbf{q}^\mathsf{T}_{j}\end{bmatrix}^\mathsf{T}$ and the fundamental matrix model $\mathbf{f}_{j}=\begin{bmatrix}\mathrm{F}^{1}_{j}&\mathrm{F}^{2}_{j}&\mathrm{F}^{3}_{j}\end{bmatrix}^\mathsf{T}$ always satisfy
\begin{equation}\label{eq::model_motion_relation}
    [\mathbf{t}_{j}]_{\times}R(\mathbf{q}_{j}) =s \mathrm{K}^\mathsf{T}
    \begin{bmatrix}
        \mathrm{F}^{1}_{j}\\
        \mathrm{F}^{2}_{j}\\
        \mathrm{F}^{3}_{j}
    \end{bmatrix}
    \mathrm{K},
\end{equation}
where $s$ is a scalar, $[\mathbf{t}_{j}]_{\times}$ is the skew-symmetric matrix representation of $\mathbf{t}_{j}$, and $R(\mathbf{q}_{j})$ is the rotation matrix of the unit quaternion vector $\mathbf{q}_{j}$.
Let $\Theta$ be the implicit function between $\mathbf{f}_{j}=\begin{bmatrix}\mathrm{F}^{1}_{j}&\mathrm{F}^{2}_{j}&\mathrm{F}^{3}_{j}\end{bmatrix}^\mathsf{T}$ and $\mathbf{p}_{j}=\begin{bmatrix}\mathbf{t}^\mathsf{T}_{j}&\mathbf{q}^\mathsf{T}_{j}\end{bmatrix}^\mathsf{T}$,
\begin{equation}\label{eq::motion_estimate}
    \Theta(\mathbf{f}_{j},\mathbf{p}_{j}) =  s\mathrm{K}^\mathsf{T}
    \begin{bmatrix}
        \mathrm{F}^{1}_{j}\\
        \mathrm{F}^{2}_{j}\\
        \mathrm{F}^{3}_{j}
    \end{bmatrix}
    \mathrm{K}
    -[\mathbf{t}_{j}]_{\times}R(\mathbf{q}_{j})
    =\mathrm{0}_{3\times3}.
\end{equation}

Ultimately, the uncertainty of $\mathbf{p}_{j}$ depends on the error distribution of model instantiation samples $\begin{bmatrix}\hdots,\mathbf{X}^\mathsf{T}_{s_{k}},\hdots\end{bmatrix}^\mathsf{T}$ and is propagated through the mapping function $\Omega^{-1}(\cdot)$ and $\Theta(\cdot)$. The noise distribution of $\mathbf{X}_{s_{k}}$ is modeled as a zero-mean Gaussian with the covariance matrix $\mathrm{\Sigma}_{X}=\sigma^{2}\mathrm{I}_{4}$ according to the assumption $\textbf{a.2}$.

Before estimating the covariance matrix of $\mathbf{f}_{j}$ to characterize its uncertainty, it is necessary to know how fundamental matrix estimation is parameterized:
\begin{itemize}
\item[] \textbf{Exact case}:
    The fundamental matrix is parameterized by the same amount of parameters as the degrees of freedom (DoFs) of fundamental matrix. For example, the 1-point algorithm is an exact case since the fundamental matrix is parameterized by one yaw angle and the DoF of fundamental matrix is $1$ when the robot motion is assumed to follow the Ackermann steering model on a planar surface.
\item[] \textbf{Overdetermined case}:
    The amount of parameters used to parameterize the fundamental matrix is larger than the DoFs of fundamental matrix. For example, the normalized 8-point algorithm is a common overdetermined case since it employs $8$ parameters but the DoFs of a general fundamental matrix is $7$.
\end{itemize}
The choice of parameterization method affects how we estimate the uncertainty.

We utilize the first order approximation~\cite{hartley2003multiple} of the covariance matrix to estimate the uncertainty. Denote the covariance matrix of $\mathbf{f}_{j}$ as $\mathrm{\Sigma}_{f_{j}}$. When the fundamental matrix estimation is the exact case, the first order approximation of $\mathrm{\Sigma}_{f_{j}}$ is
\begin{equation}
    \mathrm{\Sigma}_{f_{j}} =
        \Big(J^\mathsf{T}_{\Omega}\mathrm{\Sigma}^{-1}_{X}J_{\Omega}\Big)^{-1},
\end{equation}
where Jacobian matrix $J_{\Omega}=\frac{\partial\Omega}{\partial\mathbf{f}_{j}}$.
For the overdetermined case, let us use the $8-$point method  fundamental matrix estimation as an example. We have
\begin{equation}
    \mathrm{\Sigma}_{f_{j}} =
    \mathrm{A}\Big(\mathrm{A}^\mathsf{T}\Big(J^\mathsf{T}_{\Omega}\mathrm{\Sigma}^{-1}_{X}J_{\Omega}\Big)\mathrm{A}\Big)^{-1}\mathrm{A}^\mathsf{T},
\end{equation}
where $\mathrm{A}$ is a $9\times8$ matrix and its column vectors are perpendicular to $\mathbf{f}_{j}$. We obtain $\mathrm{A}$ by using the first $8$ columns of the Householder matrix of $\mathbf{f}_{j}$ .

The uncertainty of $\mathbf{p}_{j}$ depends on the uncertainty of $\mathbf{f}_{j}$ and $\Theta(\cdot)$. Denote the covariance matrix of $\mathbf{p}_{j}$ as $\mathrm{\Sigma}_{p_{j}}$. Under the Gaussian noise assumption, the first-order approximation of $\mathrm{\Sigma}_{p_{j}}$ is
\begin{equation}\label{eq::cov_p}
    \mathrm{\Sigma}_{p_{j}} = J_{p}\mathrm{\Sigma}_{f_{j}}J^\mathsf{T}_{p},
\end{equation}
where Jacobian matrix can be obtained either by $J_{p}=-\Big(\frac{\partial\Theta}{\partial\mathbf{p}_{j}}\Big)^{+}\frac{\partial\Theta}{\partial\mathbf{f}_{j}}$ explicitly where symbol $^{+}$ is matrix pseudo inverse, or calculating from the fundamental matrix decomposition~\cite{papadopoulo2000estimating}.

\subsubsection{Samples and $\mathbf{p}_{j}$ consistence test}\label{sec::p_comp_tests}
For the over-parameterized case, a good model must be consistent with the samples that instantiate it. Therefore, we verify if it is the case.
Only the model which passes this consistence test can advance to the next step. Otherwise, we discard the model and the $j$-th iteration ends. Of course, this does not apply to the exact case where its model always perfectly fits the samples.

The consistence between the samples and the model is measured by the error distance. We use the Sampson error vector~\cite{hartley2003multiple} to form the error measurement. Denote the Sampson error vector of $\mathbf{X}_{s_{k}}$ as $\bm{\delta}_{s_{k}}$. Let $\Delta$ as the Sampson error vector function which utilizes $\mathbf{p}_{j}$ to calculate the Sampson correction of $\mathbf{X}_{s_{k}}$
\begin{equation}\label{eq::sampson_error_vec}
    \bm{\delta}_{s_{k}} = \Delta(\mathbf{X}_{s_{k}},\mathbf{p}_{j})\in\mathbb{R}^{4}.
\end{equation}

We model $\bm{\delta}_{s_{k}}$ as the zero-mean Gaussian distribution with the covariance $\mathrm{\Sigma}_{\delta_{s_{k}}}$. Under the Gaussian noise assumption, the first-order approximation of $\mathrm{\Sigma}_{\delta_{s_{k}}}$ is
\begin{equation}\label{eq::cov_sampson_error_vec}
        \mathrm{\Sigma}_{\delta_{s_{k}}} = J_{\delta,X}\mathrm{\Sigma}_{X}J^\mathsf{T}_{\delta,X} + J_{\delta,p}\mathrm{\Sigma}_{p_{j}}J^\mathsf{T}_{\delta,p},
\end{equation}
where Jacobian matrices $J_{\delta,X} = \frac{\partial\Delta}{\partial\mathbf{X}_{s_{k}}}$ and $J_{\delta,p} = \frac{\partial\Delta}{\partial\mathbf{p}_{j}}$.

For each sample $\mathbf{X}_{s_{k}}$, we design the following hypothesis testing:
\begin{equation}
\begin{aligned}
    \mathbf{H_{0}}:\quad&\mathbf{X}_{s_{k}}\text{ does not fit }\mathbf{p}_{j},\hspace{1in}\\
    \mathbf{H_{1}}:\quad&\mbox{Otherwise}.\label{eq::sample-p-test}
\end{aligned}
\end{equation}

Given the Sampson error vector $\bm{\delta}_{s_{k}}$ (\ref{eq::sampson_error_vec}) and the covariance matrix $\mathrm{\Sigma}_{\delta_{s_{k}}}$ (\ref{eq::cov_sampson_error_vec}), the error distance is re-written as
\begin{equation}\label{eq::mahalanobis_dist_F}
    \bm{\delta}^\mathsf{T}_{s_{k}}\mathrm{\Sigma}^{-1}_{\delta_{s_{k}}}\bm{\delta}_{s_{k}}.
\end{equation}
Since we approximate $\bm{\delta}_{s_{k}}$ as the normal distribution with zero mean vector and the covariance $\mathrm{\Sigma}_{\delta_{s_{k}}}$, (\ref{eq::mahalanobis_dist_F}) is a $\chi^{2}$ distribution. Besides, $\bm{\delta}_{s_{k}}$ is defined on the variety of $\mathbf{x}^\mathsf{T}_{s_{k}}\mathrm{K}^{-\mathsf{T}}[\mathbf{t}_{j}]_{\times}R(\mathbf{q}_{j})\mathrm{K}^{-1}\mathbf{x}_{s_{k}}$, which reduces $1$ DoF. Therefore, (\ref{eq::mahalanobis_dist_F}) follows a $\chi^{2}$ distribution with $4-1=3$ DoFs. Define $F_{3}$ as the cumulative $\chi^{2}$ distribution under $3$ DoFs and we can set the distance threshold $F^{-1}_{3}(1-\alpha)$ by setting the significance level $\alpha=0.05$,
where $F^{-1}_{3}(\cdot)$ is the inverse function of $F_{3}(\cdot)$. Thus, we consider $\mathbf{X}_{s_{k}}$ agrees with $\mathbf{p}_{j}$ by rejecting $\mathbf{H_{0}}$ when
\begin{equation}
   \bm{\delta}^\mathsf{T}_{s_{k}}\mathrm{\Sigma}^{-1}_{\delta_{s_{k}}}\bm{\delta}_{s_{k}} \leq F^{-1}_{3}(1-\alpha).
\end{equation}

\subsection{Model Verification}
For the model which passes the aforementioned consistence tests, we find inliers from the rest of the inputs and verify the model by checking the quality of its inliers (See Boxes 2(a) and 2(b) in Fig.~\ref{fig::RCME_system_diagram}).

\subsubsection{Find inliers}
An inlier is defined to be a point consistent with the model. We employ $\mathbf{p}_{j}$ consistence test to find the inliers.

For the point $\mathbf{X}_{i}$, the Sampson error vector and corresponding covariance are denoted as $\bm{\delta}_{i}$ and $\mathrm{\Sigma}_{\delta_{i}}$, respectively, and can be obtained from (\ref{eq::sampson_error_vec}) and (\ref{eq::cov_sampson_error_vec}). Let $I_{j}$ as an inlier indicator function of $\mathbf{p}_{j}$
\begin{equation}\label{eq::find-inlier}
    I_{j}(\mathbf{X}_{i}) :=
    \begin{cases}
        1,\quad\text{when}\quad\bm{\delta}^\mathsf{T}_{i}\mathrm{\Sigma}^{-1}_{\delta_{i}}\bm{\delta}_{i}\leq F^{-1}_{3}(1-\alpha)\\
        0,\quad\text{otherwise}.
    \end{cases}
\end{equation}
The inlier set of $\mathbf{p}_{j}$ is $$\mathcal{X}_{j}:= \Big\{\mathbf{X}_{i_{k}}: I_{j}(\mathbf{X}_{i_{k}})=1,~~  1 \leq i_{k} \leq n \Big\}$$ with its size defined as $n_j = |\mathcal{X}_{j}|$.

\subsubsection{Score inliers} It measures the quality of the consistence between model and its inlier set. Instead of using $n_j$, the number of inliers or the loss  function~\cite{torr2000mlesac} to score the inliers, we want to score inliers by using the differential entropy on the covariance matrices $\mathrm{\Sigma}_{\delta_{i_{k}}}$. The intuition is that the joint distribution of distances $\{ \bm{\delta}_{i_{k}}, k=1,2,...,n_j\}$ for the inliers should have a small entropy for a high quality inlier set. For inlier set $\mathcal{X}_{j}$, we define a score vector $\mathbf{h}_{j}$ as,
\begin{equation}\label{eq::score-inliers}
    \mathbf{h}_{j} = [\hdots, h_{i_{k},j}, \hdots ]^\mathsf{T}\in\mathbb{R}^{n_j},
\end{equation}
where each entry is a differential entropy for each inlier,
\begin{equation}
    h_{i_k,j} = \frac{1}{2}\log((2\pi)^{4}\exp(4)|\mathrm{\Sigma}_{\delta_{i_{k}}}|)
\end{equation}
where $|\mathrm{\Sigma}_{\delta_{i_{k}}}|$ denotes the determinant of $\mathrm{\Sigma}_{\delta_{i_{k}}}$.

\subsubsection{Inlier set quality test}\label{sec::inlier_qual_test}
Now, we evaluate the inlier's quality to determine if the model can enter candidate solution set by checking entropy values. A good model must contain the inliers with high quality of consistence and leads to small entropy values.  Given the score vector $\mathbf{h}_{j}$, the average entropy and the standard deviation are defined as follows:
\begin{equation}
    \psi_{j} = \frac{\|\mathbf{h}_{j}\|_{1}}{n_j}, ~~
    s_{j} = \sqrt{\frac{\|\mathbf{h}_{j}-\psi_{j}\mathbf{1}_{n}\|^{2}_{2}}{n_j-1}},
\end{equation}
where $\|\cdot\|_{1}$ is L$1$ norm, $\|\cdot\|_{2}$ is L$2$ norm, and $\mathbf{1}_{n}$ is a n-vector of ones.

To evaluate the quality of the model and its inlier set, we design the following hypothesis testing based on the $Z-$test,
\begin{equation}
\begin{aligned}\label{eq::inlier-quality-test}
    \mathbf{H_{0}}:\quad&\psi_{j}>\mu,\hspace{1.5in}\\
    \mathbf{H_{1}}:\quad&\mbox{Otherwise},
\end{aligned}
\end{equation}
where $\mu=-3.53$ is an differential entropy threshold determined by the experiments. The test statistic can be calculated
\begin{equation}
    Z_{j} = \frac{\psi_{j} - \mu}{s_{j}/\sqrt{n_j}}.
\end{equation}
Define $\Phi(x)$ as the cumulative distribution function of the standard normal distribution at value $x$. By setting the significance level $\alpha$, the p-value is obtained $\Phi^{-1}(1-\alpha).$
We consider that the model is highly consistent with its inlier set by rejecting $\mathbf{H_{0}}$ when
\begin{equation}
    Z_{j} \leq \Phi^{-1}(1-\alpha).
\end{equation}
For models that passed the hypothesis testing, we proceed to next step.

A good model must contain both sufficiently large amount of and high quality inliers. We use two ratio thresholds $\omega_{p}$ and $\lambda$ to determine if the $j$-th model satisfies the requirement. $\omega_{p}$ is a priorly-known or estimated inlier ratio. A model is considered to contain a sufficiently large amount of inliers when
\begin{equation}\label{eq:inlier_size_cond}
    \frac{n_j}{n}\geq \lambda \omega_{p},
\end{equation}
where conservative coefficient $0.5 \leq \lambda \leq 1$ determine how close to ideal size  $\omega_{p}$ we want the inlier set to be. Same as the RANSAC, inlier ratio $\omega_{p}$ can be inferred in the process. All models that survive the hypothesis testing in (\ref{eq:inlier-test}) and size condition in (\ref{eq:inlier_size_cond}) are added to the candidate solution set,
\begin{equation}\label{eq::candidate-set}
\mathcal{C} := \left\{j | \bigl( Z_{j} \leq \Phi^{-1}(1-\alpha)\bigr) ~ \land ~ \Bigl(\frac{n_j}{n}\geq \lambda \omega_{p}\Bigr), \forall j\right\}.
\end{equation}
After the maximum number of trials $N$, we consider the camera motion recovery fail if $\mathcal{C} = \emptyset$. This leads to the failed algorithm output (i.e. `$\star$' in Fig.~\ref{fig::RCME_system_diagram}). The failure reason can be either that the iteration number is not big enough or that the input $\{\mathbf{X}_{i}\}$ is from poor quality key frames.

\subsection{Model Selection and Early Termination}
\subsubsection{Model selection} For non empty $\mathcal{C}$, we select the one with the minimum average entropy to be the output.
\begin{equation}\label{eq::model_select_entropy}
    \min_{j\in\mathcal{C}}\frac{\|\mathbf{h}_{j}\|_{1}}{n_j}.
\end{equation}
This output serves as an initial solution for the following model refinement by applying Maximum Likelihood Estimation (MLE) to minimize reprojection error. Since this is the same as the traditional approach (i.e. Box 4s in both Fig.~\ref{fig::RANSAC_system_diagram} and Fig.~\ref{fig::RCME_system_diagram}, we skip it here.

\subsubsection{Early termination condition}\label{sec::early-terminate}
So far, our RCME algorithm runs for the entire maximum iterations. It is possible to design an early termination threshold to speed up the algorithm. Note that entries in $\mathcal{C}$ grow after each successful iteration. For each new entry, we can test its average entropy $\frac{\|\mathbf{h}_{j}\|_{1}}{n_j}$ by comparing to a preset threshold. If it is small enough, which means it is a satisfying solution, we can terminate RCME early to perform the MLE-based model refinement.

\section{Algorithm and Complexity Analysis}

\begin{algorithm}[ht]\label{alg::RCME}
    \SetKwInOut{Input}{Input}
	\SetKwInOut{Output}{Output}
    \caption{RCME algorithm}
	\Input{$\mathbf{X}_{1},\cdots,\mathbf{X}_{n}$}
	\Output{Fail or \{$\mathbf{q}$, $\mathbf{t}$\}}
    Initialize maximum iteration number $N$; \hfill $O(1)$\\
    Initialize the model ranking $\mathcal{C}=\emptyset$; \hfill $O(1)$\\
    \For{$j=1$ to $N$}{
        Select minimal sample set $\mathcal{S}_{j}$; \hfill $O(1)$\\
        Estimate $\mathbf{f}_{j}$ and $\mathbf{p}_{j}$ using (\ref{eq::model_instantiate}) and (\ref{eq::motion_estimate}); \hfill $O(1)$\\
        \If{$\mathbf{f}_{j}$ is overdetermined}{
            Perform $\mathcal{S}_{j}$ and $\mathbf{p}_{j}$ consistence test (\ref{eq::sample-p-test}); \hfill $O(1)$\\
            \If{$\mathcal{S}_{j}$ fails in consistence tests}{
                continue\
            }
        }
        Find and score inliers $\mathbf{h}_{j}$ using (\ref{eq::find-inlier}) and (\ref{eq::score-inliers}); \hfill $O(n)$\\
        Perform inlier quality tests (\ref{eq::inlier-quality-test}); \hfill $O(n)$\\
        \If{$\mathbf{p}_{j}$ fails in inlier quality test}{
            continue\
        }
        Update the candidate set $\mathcal{C}$ by (\ref{eq::candidate-set}); \hfill $O(1)$\\
        \If{$\mathbf{h}_{j}$ satisfies early termination (Sec.~\ref{sec::early-terminate})}{
            break\
        }
    }
    \If{$\mathcal{C}=\emptyset$}{
        \Return Fail
    }
    Select $\mathbf{p}$ from $\mathcal{C}$ using (\ref{eq::model_select_entropy}); \hfill $O(N)$\\
    Perform MLE on $\mathbf{p}$; \hfill $O(n^{3}/\epsilon^{2})$\\
    \Return $\mathbf{q}$, $\mathbf{t}$
\end{algorithm}

Our RCME algorithm is summarized in Algorithm~\ref{alg::RCME} to facilitate our complexity analysis. Each iteration begins with the model instantiation and it takes $O(1)$ to select samples from $n$ points by using pseudo-random number generator algorithm~\cite{vigna2016experimental}. Model estimation takes $O(1)$ time. Performing the model consistence test on every sample takes $O(1)$ since $m$ is a small and fixed constant. It takes $O(n)$ time to determine whether each point $\mathbf{X}_{i}$ is an inlier or not. Updating the model candidate $\mathcal{C}$ takes constant time to check if the model passes the inlier quality test and satisfies the size condition. The overall computational complexity for each iteration is $O(n)$. The total iteration performed in RCME is bounded below $O(N)$. Thus, the overall iteration process takes $O(Nn)$.

Suppose on average $N_{C}$ models enter the final candidate set, where $N_{C}< N$. Therefore, model selection takes $O(N)$. MLE is solved by applying the Levenberg-Marquardt (LM) algorithm~\cite{hartley2003multiple}. In each iteration of LM, the dense matrix solver takes $O(n^{3})$. The total iteration needed by LM is bounded below $O(1/\varepsilon^{2})$ with the stopping threshold $\varepsilon$.
The overall computational complexity for $\mathbf{p}$ re-estimation is $O(n^{3}/\varepsilon^{2})$. To summarize the analysis, we have the following,
\begin{Thm}
    The computational complexity of our RCME algorithm is $O(Nn + n^{3}/\varepsilon^{2})$.
\end{Thm}

\section{EXPERIMENTS}

\subsection{Algorithms in Comparison and Settings}
We have implemented our system in C++. We evaluate algorithm robustness using frequently-used public data sets. We compare RCME with the Gold standard approach with RANSAC for computing the fundamental matrix~\cite{hartley2003multiple} which is the most widely-used fundamental matrix estimation method. We abbreviate the the Gold standard approach as ``Standard'' in the comparison. We also include ``pRCME'' which is RCME with model consistence test (\ref{eq::sample-p-test}) in Sec.~\ref{sec::p_comp_tests} turned off. The purpose is to show the individual effectiveness of the consistence test and the following inlier quality test.

All algorithms employ the normalized $8$-point algorithm~\cite{hartley2003multiple} for the fundamental matrix estimation since it is fast and works for general camera motion. The maximum iteration number $N$ is set to $200$. We perform a complete $200$ iterations instead of estimating $N$ during the iteration process. This follows the setup in ORB-SLAM2~\cite{mur2017orb}. The noise $\sigma$ is set to $0.5$.

\subsection{Testing Datasets} We test our method with a wide range of public datasets: KITTI odometry dataset~\cite{geiger2013vision}, EuRoC MAV dataset~\cite{burri2016euroc}, and HRBB 4th floor dataset in Texas A$\&$M University (HRBB4)~\cite{yan-mfg-lba-tro-2015}. We divide test datasets into two groups: indoor datasets and outdoor datasets because their feature distributions are quite different.

Indoor datasets include HRBB4 and EuRoC MAV. The datasets cover different indoor environments. The HRBB4 dataset is recorded in the office corridor environment. The EuRoC MAV dataset is collected by a synchronized stereo camera. Since this paper only concerns monocular vision, we only utilize the images from the left camera. The EuRoC MAV dataset contains two types of indoor environments, one is recorded in a machine hall and the other is recorded in a vicon room. The EuRoC MAV dataset has $11$ sequences with different types of camera motion and lighting conditions. It covers from office rooms to industrial halls with different objects and camera motions.

The outdoor dataset includes KITTI odometry dataset. The KITTI odometry dataset is recorded by a moving vehicle while driving in Karlsruhe, Germany. We perform tests on $11$ image sequences of the KITTI odometry dataset.

In all cases, the sequence of two-view image pairs of each dataset are initially selected by applying the initialization module (i.e. key frame selection) of ORB-SLAM2~\cite{mur2017orb} and we only keep the non-degenerated two-view image pairs. For each two-view image pair, we use SIFT~\cite{lowe2004distinctive} to obtain the point correspondences as our inputs.

\subsection{Evaluation Metric for Robustness} \label{sec::metric}
Robustness is measured by failure rate of each algorithm. It is important to recognize a failed case.

For every two-view pair, we use the ratio of the consistent inlier amount before and after performing MLE in model refinement step (Box 4s in Fig.~\ref{fig::RANSAC_system_diagram} and Fig.~\ref{fig::RCME_system_diagram}) to identify if the method fails to find a correct solution. Let $n_{I}$ and $n^{*}_{I}$ be the consistent inlier amount before and after MLE, respectively.
A quality solution should cause the number of inliers to increase or at least maintain its inlier set size instead of decreasing inlier set size drastically. The underlying rationale is that an incorrect solution usually falls into local minima instead of the solution close to the global minima in the optimization process. The local minima causes the solution loses the consistence from its inlier data.

We employ the Huber robust function $\gamma(\cdot)$ on top of reprojection error (page 617 in~\cite{hartley2003multiple} 
) to re-evaluate the consistence of inliers before and after MLE to get $n_{I}$ and $n^{*}_{I}$, respectively. An inlier remains consistent with its model when the value of the Huber robust function is less than the threshold $\tau^{2}=\sigma^{2}F^{-1}_{2}(1-\alpha)$, where $F_{2}$ denotes the cumulative $\chi^{2}$ distribution under $2$ DoF and $\alpha=0.05$ is the significance level.

We set a ratio threshold $\kappa=0.5$ to determine that the solution is considered as a failed camera motion recovery when
\begin{equation}\label{eq::metric}
    \frac{n^{*}_{I}}{n_{I}} \leq \kappa.
\end{equation}
We collect the overall failure rate for each dataset as the measure for robustness for the dataset.

Note that precision that is represented by residual error in cost function is not the concern here because both standard RANSAC and RCME can output high quality solutions in precision if they do NOT fail. In other words, precision is useless if they fail to find the correct solution.

It is worth noting that we do not use the provided ``ground truth" from some datasets because those ``ground truth" were mostly computed from the Gold Standard approach and its correctness is not well verified. Only a small ratio of wrong ground truth would cause comparison issues.

\subsection{Experimental results}

The experimental results are shown in Tab.~\ref{table::real-data-test}. The upper half of the table are results from outdoor datasets and the lower half of table are results from indoor datasets.

For both RCME and pRCME, they have the ability to detect poor quality inputs. In those case, the algorithms do not proceed and directly output failure which are represented as ``Detect-fail'' columns. The ``Failure'' columns represent the actual failure computed using the metric in Sec.~\ref{sec::metric}. The ``Failure'' columns do not include cases in ``Detect-fail'' columns.
In general, we want the failure rate to be as close to zero as possible. It is clear that both RCME and pRCME outperform the ``Standard'' approach, even you may add ``Detect-fail'' column with ``Failure'' column together and compare the values to those of the ``Standard'' approach. Our algorithms have improved robustness of RANSAC for camera motion estimation.

It is also clear that RCME is better than pRCME, which indicates that our model consistence test with its samples works as expected. Also, the fact that pRCME is better than the ``Standard'' approach means that our inlier-quality test works as expected. It is worth noting that the performance improvement is more significant for indoor cases. This is true because indoor datasets often suffer more from uneven feature distribution and poor light conditions. This means RCME handles those challenging conditions better. More specifically, the average failure rate for indoor datasets is reduced from $1.41\%$ to $0.02\%$.

\begin{table}[ht!]
	\centering
    \caption{Experimental Results. Best results are highlighted in boldface.}
	\label{table::real-data-test}
    \scalebox{0.65}{
	\begin{tabular}{  l c c c c c}
		\toprule[0.8pt]
         & \multicolumn{2}{ c }{RCME} & \multicolumn{2}{ c }{pRCME} & Standard\\
        \cline{2-6}
        Dataset & Detect-fail \% & Failure \% & Detect-fail \% & Failure \%  & Failure \% \\ \hline
        KITTI/00(\#2412) & 0.00 & \textbf{0.00} & 0.00 & \textbf{0.00} & \textbf{0.00}\\
        KITTI/01(\#56) & 0.00 & \textbf{0.00} & 0.00 & \textbf{0.00} & 1.79\\
        KITTI/02(\#2031) & 0.00 & \textbf{0.00} & 0.00 & 0.05 & 0.05\\
        KITTI/03(\#358) & 0.00 & \textbf{0.00} & 0.00 & \textbf{0.00} & \textbf{0.00}\\
        KITTI/04(\#86) & 0.00 & \textbf{0.00} & 0.00 & \textbf{0.00}  & \textbf{0.00}\\
        KITTI/05(\#1489) & 0.00 & \textbf{0.00} & 0.00 & \textbf{0.00} & 0.27 \\
        KITTI/06(\#295) & 0.00 & \textbf{0.00} & 0.00 & \textbf{0.00} & \textbf{0.00}\\
        KITTI/07(\#564) & 0.00 & \textbf{0.00} & 0.00 & \textbf{0.00} & \textbf{0.00}\\
        KITTI/08(\#1690) & 0.00 & \textbf{0.00} & 0.00 & \textbf{0.00} & 0.18\\
        KITTI/09(\#678) & 0.00 & \textbf{0.00} & 0.00 & 0.15 & \textbf{0.00} \\
        KITTI/10(\#419) & 0.00 & \textbf{0.00} & 0.00 & \textbf{0.00} & 0.24\\ \hline
        Avg. & 0.00 & \textbf{0.00} & 0.00 & 0.02 & 0.10\\ \hline
        EuRoC/MH\_01(\#515) & 0.19 & \textbf{0.00} & 0.00 & 0.19          & 0.19\\
        EuRoC/MH\_02(\#450) & 0.00 & \textbf{0.00} & 0.00 & \textbf{0.00} & 0.89  \\
        EuRoC/MH\_03(\#587) & 0.00 & \textbf{0.00} & 0.00 & \textbf{0.00} &\textbf{0.00}  \\
        EuRoC/MH\_04(\#357) & 0.28 & \textbf{0.00} & 0.28 & \textbf{0.00} & 0.84  \\
        EuRoC/MH\_05(\#393) & 0.25 & \textbf{0.00} & 0.00 & \textbf{0.00} & 0.25 \\
        EuRoC/V1\_01(\#566) & 0.00 & \textbf{0.00} & 0.00 & \textbf{0.00} & 1.77  \\
        EuRoC/V1\_02(\#383) & 0.26 & \textbf{0.00} & 0.26 & \textbf{0.00} & 1.83  \\
        EuRoC/V1\_03(\#108) & 0.93 & \textbf{0.00} & 0.93 & \textbf{0.00} & 8.33  \\
        EuRoC/V2\_01(\#391) & 0.26 & \textbf{0.00} & 0.26 & 0.26          & 2.81  \\
        EuRoC/V2\_02(\#590) & 0.00 & \textbf{0.17} & 0.00 & \textbf{0.17} & 3.05 \\
        EuRoC/V2\_03(\#92)  & 0.00 & \textbf{0.00} & 0.00 & \textbf{0.00} & 3.26 \\
        HRBB4th(\#306)      & 0.00 & \textbf{0.00} & 0.00 & \textbf{0.00} & 0.33 \\
        \hline
        Avg. & 0.13 & \textbf{0.02} & 0.08 & 0.06 & 1.41\\
        \toprule[0.8pt]
	\end{tabular}
    }
	\centering
\end{table}

\section{CONCLUSIONS AND FUTURE WORK}

We reported our new robust camera motion estimation algorithm targeted at improving robustness in traditional RANSAC-based approaches. Combining two new developments: model quality test with its samples, and inlier quality test with its model, we are able to consistently reduce failure rate of the existing algorithm, as shown in the experimental results from testing a wide range of indoor and outdoor datasets.

In the future, we will look deep into those failed cases that have not been detected by our algorithm and try to design new tests to further improve the robustness of the entire algorithm. We are in the processing of embedding our estimator into existing open source implementation such as ORB-SLAM2 to share the new developments with others.

{\small
\section*{Acknowledgment}
We thank Y. Xu for his insightful discussion. We are also grateful to H. Cheng, B. Li, A. Kingery, A. Angert, D. Wang, and Y. Ou for their inputs and contributions to the Networked Robots Laboratory at Texas A\&M University.

\bibliographystyle{plain}
\bibliography{Yeh,dez,Yan}
}

\end{document}